\documentclass{article}

\usepackage[preprint]{neurips_2025}

\usepackage[numbers]{natbib}     

\usepackage[utf8]{inputenc} 
\usepackage[T1]{fontenc}    
\usepackage{hyperref}       
\usepackage{url}            
\usepackage{booktabs}       
\usepackage{amsfonts}       
\usepackage{nicefrac}       
\usepackage{microtype}      
\usepackage{xcolor}         

\usepackage{amsmath}        
\usepackage{graphicx}       
\usepackage{multirow}
\usepackage{makecell}

\usepackage{graphicx} 
\usepackage{lipsum}   

\title{LayoutRAG: Retrieval-Augmented Model for Content-agnostic Conditional Layout Generation}

\author{%
  Yuxuan Wu \\
  Xi'an Jiaotong University \\
  \texttt{yuxuan9862@gmail.com} \\
  \And
  Le Wang \thanks{Corresponding author} \\
  Xi'an Jiaotong University \\
  \texttt{lewang@xjtu.edu.cn} \\
  \And
  Sanping Zhou \\
  Xi'an Jiaotong University \\
  \texttt{spzhou@xjtu.edu.cn} \\
  \AND
  Mengnan Liu \\
  Xi'an Jiaotong University \\
  \texttt{Liumn@stu.xjtu.edu.cn} \\
  \And
  Gang Hua \\
  Amazon.com, Inc. \\ 
  \texttt{ganghua@gmail.com}
  \And
  Haoxiang Li \\
  Pixocial Technology \\
  \texttt{lhxustcer@gmail.com} \\
}

\begin{document}

\maketitle

\begin{abstract}
  Controllable layout generation aims to create plausible visual arrangements of element bounding boxes within a graphic design according to certain optional constraints, such as the type or position of a specific component. 
  While recent diffusion or flow-matching models have achieved considerable advances in multifarious conditional generation tasks, 
  there remains considerable room for generating optimal arrangements under given conditions.
  In this work, we propose to carry out layout generation through retrieving by conditions and reference-guided generation.
  Specifically, we retrieve appropriate layout templates according to given conditions as references. The references are then utilized to guide the denoising or flow-based transport process.  
  By retrieving layouts compatible with the given conditions, we can uncover the potential information not explicitly provided in the given condition.
  Such an approach offers more effective guidance to the model during the generation process, in contrast to previous models that feed the condition to the model and let the model infer the unprovided layout attributes directly. 
  Meanwhile, we design a condition-modulated attention that selectively absorbs retrieval knowledge, adapting to the difference between retrieved templates and given conditions.
  Extensive experiment results show that our method successfully produces high-quality layouts that meet the given conditions and outperforms existing state-of-the-art models.
  Code will be released upon acceptance. 
\end{abstract}

\section{Introduction}


Layout generation refers to creating the arrangement of various visual components, such as images, text, or other components on a canvas or document page.
A well-structured layout enables users to easily comprehend and interact effectively with the displayed information.    
The ability to generate high-quality layouts facilitates various applications like user interfaces~\cite{deka2017rico} and graphic design~\cite{zhong2019publaynet}.

In many real-world cases, we'd like the model to interact with users and provide controllable layout generation according to user specifications.
Given the subjectivity of what is a good layout, it is much more user-friendly to model the layout generation an interactive process rather than a one-off event. Technically, user interactions are considered as constraints of the generation process so that users can specify attributes like element types, positions, and sizes to guide the layout generation.
Hence, developing models that can better understand user-specified conditions and generate appropriate layouts tailored to user preferences is essential for broader applications of layout generation models.

\begin{figure}
  \centering
  \label{intro}
  \includegraphics[width=1.0\textwidth]{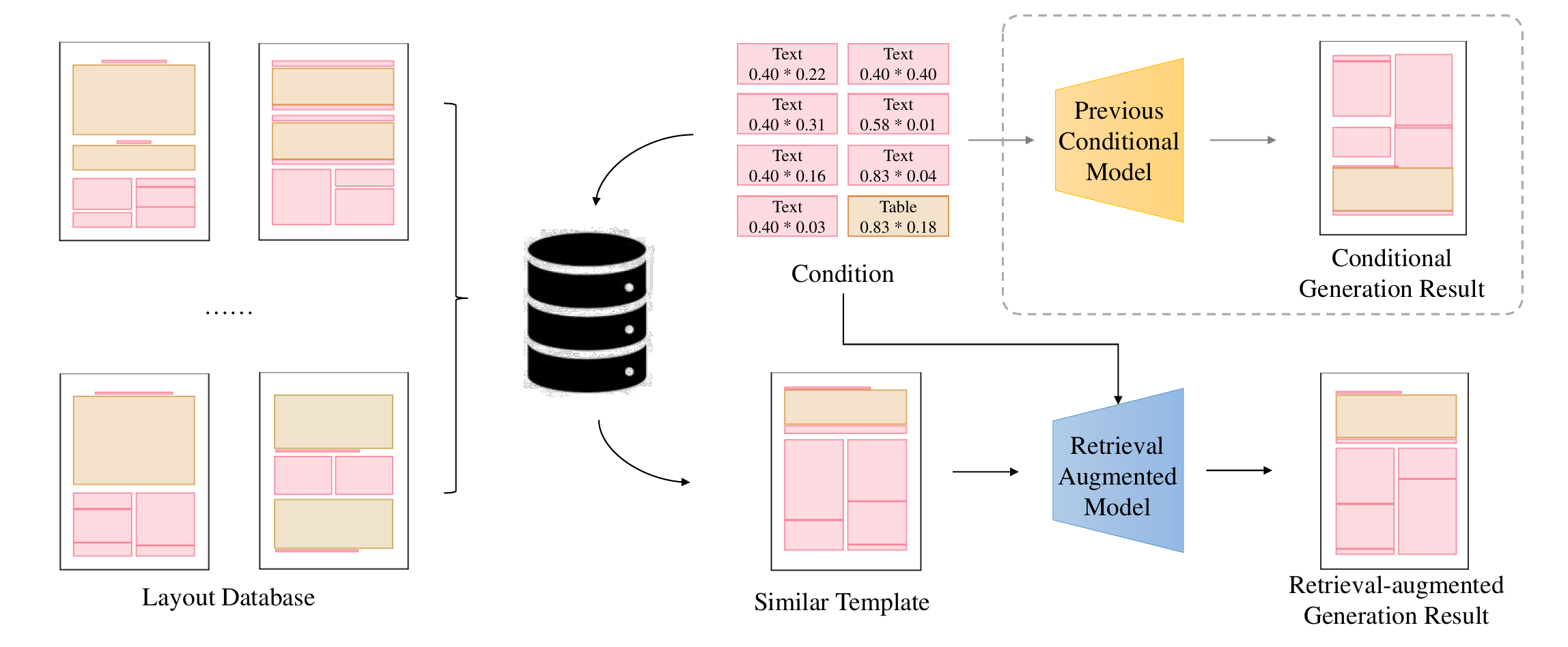}
  \caption{Conditioning mechanism of previous models (top right) and LayoutRAG.
  Benefiting from the extra knowledge from the retrieved samples, LayoutRAG can generate better conditional layouts.}
  \vspace{-0.3cm}
\end{figure}

In recent years, diffusion models and flow matching have gained much attention
in layout generation, as they have shown great promise in terms of faithfully learning a given data distribution and sampling from it.
Besides, the versatility of such models in conditional generation makes them a dominant paradigm for controllable layout generation.
LayoutDM~\cite{inoue2023layoutdm} and LayoutDiffusion~\cite{zhang2023layoutdiffusion} train an unconditional model and impose the conditions during sampling through a guidance strategy.
LayoutFormer++~\cite{jiang2023layoutformer++} takes serialized conditions and generates layouts in an autoregressive manner.
DLT~\cite{levi2023dlt}, LACE~\cite{chen2024towards}, and LayoutFlow~\cite{guerreiro2024layoutflow} feed condition masks to explicitly inform the conditioning attributes besides injecting conditions during sampling.
Although these models have achieved promising results, they may still not be able to find optimal arrangements under given conditions.
The generation result may still have undesirable overlaps or empty spaces in certain scenarios, as shown in top right of Figure ~\ref{intro}.
There remains considerable room for improvement.
Moreover, it is difficult to bias the behavior of the generation model to address failure cases without retraining it on an updated dataset with a substantial number of additional samples. 

To address these limitations, we approach conditional layout generation through a combination of condition-based retrieval and reference-guided generation, as shown in Figure ~\ref{intro}. 
Given certain conditions representing any subset of the attributes of the layout components provided (class, size and location), the model first searches for layouts compatible with the given condition in the database. 
The retrieved layouts that meet the condition or differ slightly are then selected as the final layout output, possibly with minor modifications.
If simple modification cannot produce desired results, they will be used as the reference to guide the denoising or flow-based transport process through condition-modulated attention.
This approach enables the conditions to be enriched with condition-relevant references from the dataset, extending beyond just the conditioned attributes and maximizing the use of existing layout datasets.
Additionally, the reference-guided mechanism helps compensate for the deficiencies of guidance in existing layout diffusion or flow-matching models. 
Furthermore, our proposed approach enables the model during inference to generalize to new knowledge in form of alternative layout databases without requiring further training, what can be interpreted as a form of post-hoc model modification, providing the model with more flexibility and generalization ability. 

We evaluate our method on two large-scale datasets Rico~\cite{deka2017rico} and PubLayNet~\cite{zhong2019publaynet} and observe improved performance compared to state-of-the-art methods.
We further conduct extensive ablative experiments to show the significant impact of the design choices in our model.

Overall, the main contributions of our work are summarized as follows:

\begin{itemize}
\item 
We first introduce RAG to the layout generation problem, decoupling the controllable generation process into two stages: reference layout retrieval and reference-guided generation. This two-stage process greatly improves the 
quality of layout generation in accordance with conditions
and makes it highly flexible for the model to address failure cases.

\item To provide reasonable guidance from the reference during the generation process, the Condition Modulated Attention (CMA) module is designed to selectively absorb retrieval knowledge and adapt to the difference between retrieved templates and the given conditions.

\item Extensive experiments show that our proposed method achieves comparable or better results compared to state-of-the-art models. 
Retrieval as reference-guidance can help generate better layouts and provide more flexibility to generalize to new knowledge via alternative layout databases.

\end{itemize}

\section{Related work}

\subsection{Diffusion based Layout Generation}
In recent years, diffusion models have dominated in various generation and editing tasks~\cite{song2019generative, meng2022sdedit, song2021denoising, lugmayr2022repaint, pmlr-v162-nichol22a}.
The studies of layout generation have also shifted toward using diffusion models for better generative quality and versatility in conditional generation. 
LayoutDM~\cite{inoue2023layoutdm} and LDGM~\cite{hui2023unifying} use discrete diffusion models~\cite{peng2022d3pm} to represent categorical data, and continuous coordinates are quantized into different states based on the element type. Additionally, an attribute-specific corruption strategy~\cite{gu2022vector} restricts variables to their respective sample spaces.
LayoutDiffusion~\cite{zhang2023layoutdiffusion} extends discrete diffusion with a new mild forward process closer to the continuous process.
As discrete tokens do not fit well with continuous geometry data, 
continuous diffusion is also used by several models. 
Another LayoutDM~\cite{chai2023layoutdm} uses continuous diffusion to generate continuous coordinates given categorical type. 
DLT~\cite{levi2023dlt} proposes joint discrete-continuous diffusion to generate discrete types and continuous coordinates simultaneously. LACE~\cite{chen2024towards} proposes a unified model to generate both geometric and categorical attributes for various tasks in a continuous space.
Other than diffusion, flow matching~\cite{lipman2023flow, liu2023flow} has also been introduced as a powerful generative framework. LayoutFlow~\cite{guerreiro2024layoutflow} applies flow-matching to layout generation and shows that flows offer a more intuitive generation process from a geometrical interpretation with less inference steps compared to diffusion models.


\subsection{Conditioning Mechanism}
To make the model applicable to real-world graphic design applications requiring user interaction, various controllable layout generation models have been proposed. 
Before diffusion models, conditioning mechanisms have been explored by GAN-based models ~\cite{kikuchi2021constrained, li2019layoutgan} and CVAE-based models ~\cite{lee2020neural, arroyo2021variational}.
Besides, BLT ~\cite{kong2022blt} achieves conditioning by masking the conditioned component attributes.
LayoutFormer++~\cite{jiang2023layoutformer++} takes serialized conditions and generates layouts autoregressivily.   
The recently dominant diffusion model, in addition to its great generalization ability, offers more flexible options in conditional generation.
LayoutDM~\cite{inoue2023layoutdm} and LayoutDiffusion~\cite{zhang2023layoutdiffusion} train an unconditional model and impose the conditions during sampling through a guidance strategy.
DLT~\cite{levi2023dlt}, LACE~\cite{chen2024towards}, and LayoutFlow~\cite{guerreiro2024layoutflow} feed condition masks into the model, explicitly informing the conditioned attributes besides injecting conditions during sampling.
Although these works have shown promising ability in conditional generation, 
they may still not be able to generate optimal layouts under given conditions in certain scenarios
Instead of directly inferring the unknown attributes with given conditions, our model retrieves layout templates in terms of given conditions to find the potential arrangements of the unknown attributes, providing more informative guidance for layout generation.

\begin{figure}
  \centering
  \includegraphics[width=1\textwidth]{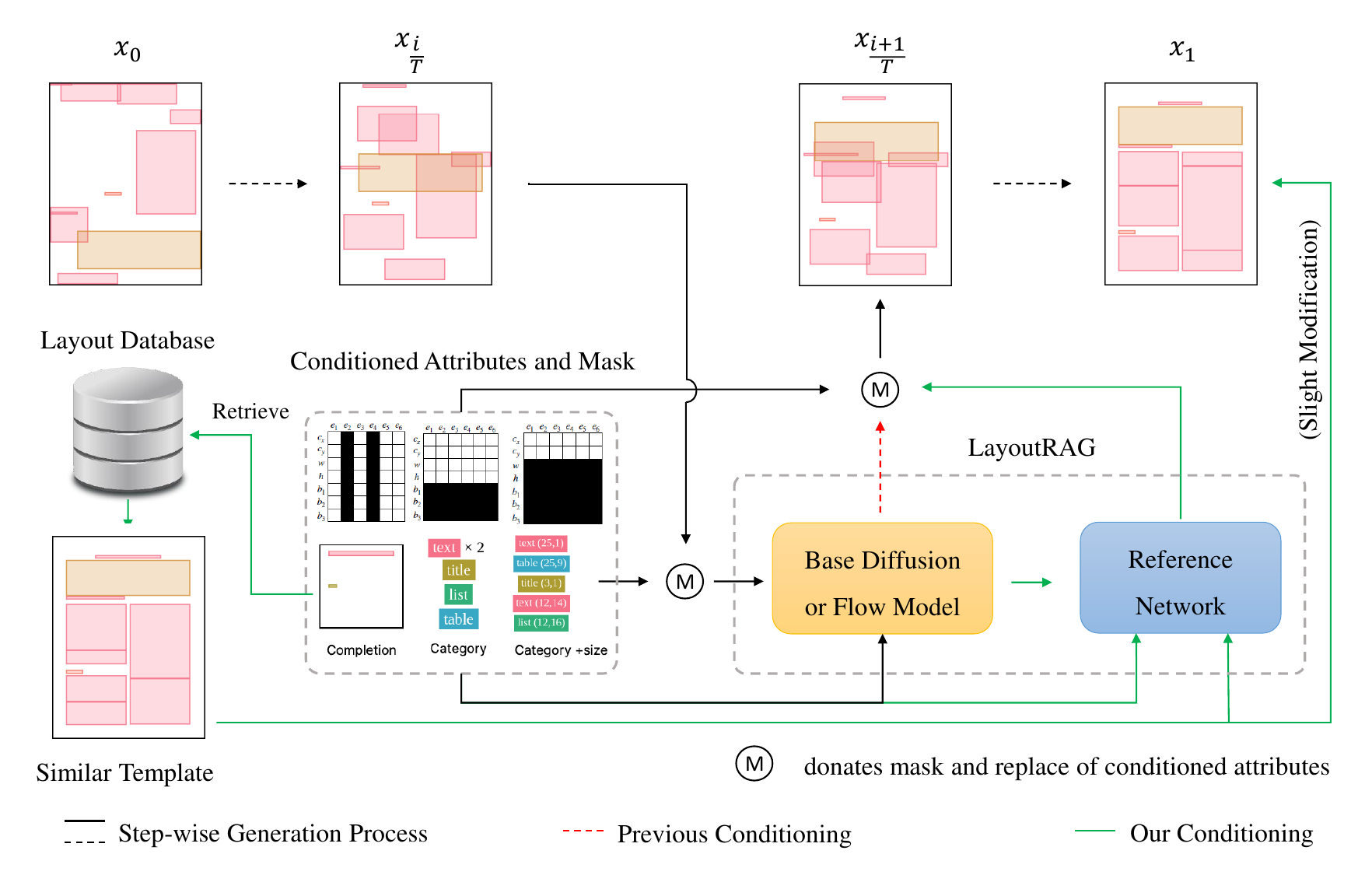}
  \caption{Overview of our retrieval-agumented layout generation model. 
  In the generation process, our model retrieves layouts by conditions.
  Retrieval layouts can be used as final results with or without slignt modifications, or be fed into reference network to guide the generation process.
  }
  \label{Main Framework Figure}
  \vspace{-0.2cm}
\end{figure}

\subsection{Retrieval Augmented Generation}
The retrieval-augmented mechanism is a widely used technique in enhancing generative models.
It enables the model to store extensive knowledge in external memory without remarkably increasing model parameters and reference informative samples to aid generation.  
In NLP, classic research has shown how retrieval-augmented methods can significantly improve the quality of generated text~\cite{guu2020retrieval, borgeaud2022improving}. Similarly, in image generation, retrieval-augmented models focus on using samples from a database to produce more realistic and high-quality images~\cite{tseng2020retrievegan, siddiqui2021retrievalfuse, xu2021texture, casanova2021instance}.
Moreover, retrieval augmentation offers additional benefits, such as
model lightweighting in RDM~\cite{blattmann2022retrieval}, generating out-of-distribution images in KNN-Diffusion~\cite{sheynin2022knn} and enhancing diffusion inference efficiency in ReDi~\cite{zhang2023redi}.
Beyond NLP and image synthesis, retrieval-augmentation has also been applied to other diversified tasks like text-driven motion sequence generation~\cite{zhang2023remodiffuse}, time series forecasting~\cite{liu2024retrieval} and embodied motion planning~\cite{oba2024read}.
In layout generation, retrieval augmentation is rarely utilized. RATF~\cite{horita2024retrieval} retrieves layout examples based on canvas image to address the data scarcity problem in content-aware scenarios, ignoring the conditions of partially known layout attributes.
However, in content-agonistic setting, canvas image retrieval is not applicable. 
To the best of our knowledge, this is the first work to explore retrieval based on partially known layout attributes in a content-agnostic layout generation setting. We observe that by presenting the model potential arrangements of unknown layout attributes, it can generate better conditional layouts.

\section{Method}

The overall framework is shown in Figure~\ref{Main Framework Figure}.
Our pipeline is based on LayoutFlow~\cite{guerreiro2024layoutflow}, 
which uses flow-matching to build a unified model for unconditional and conditional layout generation. 


\subsection{Layout Retrieval}
\label{Layout Retrieval}
\begin{figure}
  \centering
  \includegraphics[width=1.0\textwidth]{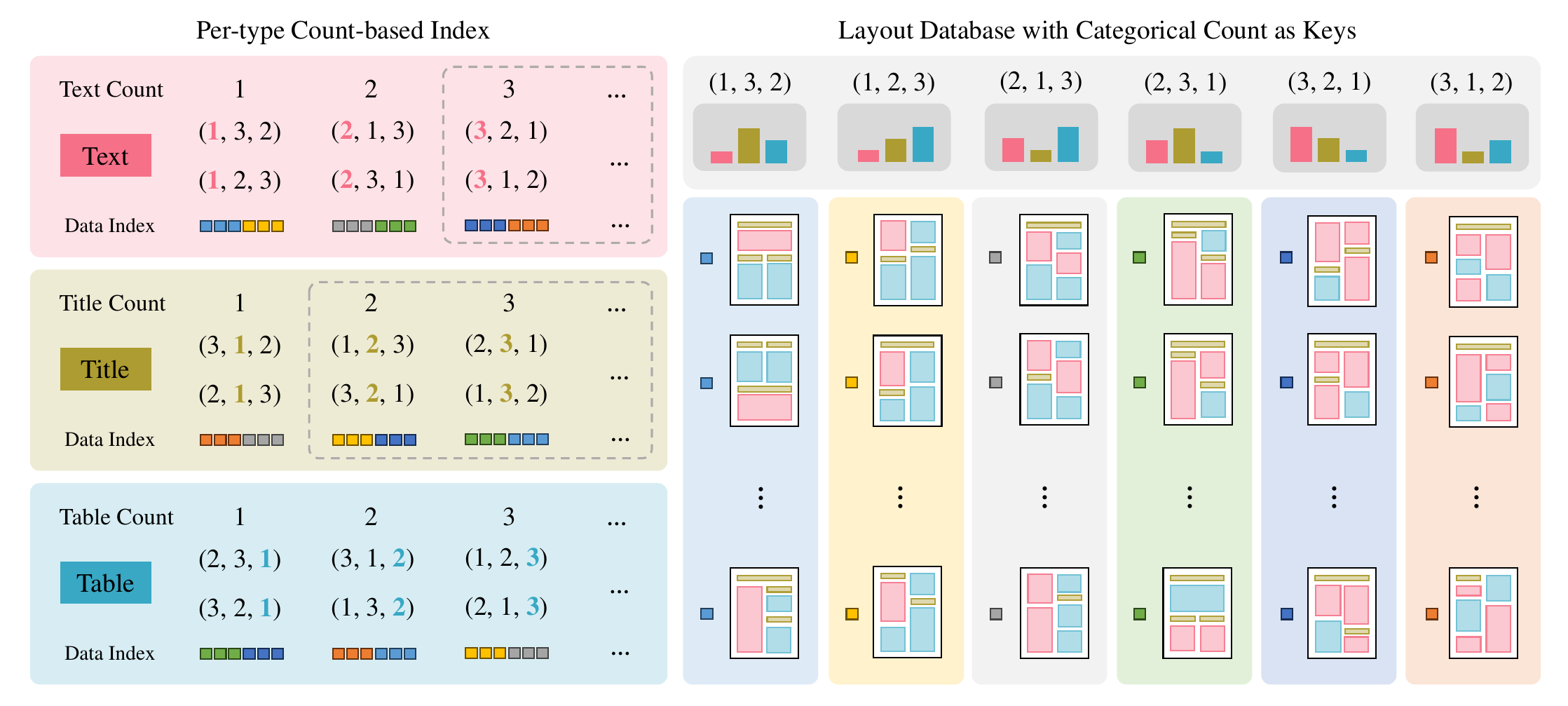}
  \caption{Overview of category count-based index system. 
  The small polychrome squares are the corresponding indices in the database.
  The top right row shows the keys formed by the count of elements for each type. 
  The left side shows the index for finding layouts with a given element count of one category, which can be used to gather layouts with a minimum number of elements in the specified category. 
  i.e. If we know the layout to be generated has at least 3 text blocks and 2 titles, the qualified layout indices for text and title are marked inside the grey dashed boxes respectively. By performing intersections between them, we can obtain the type-qualified candidates.
  }
  \label{Memory Architecture}
\end{figure}

To establish the retrieval database, we simply select all the training data as entities. 
As the volume of the layout database can be very large and the layout attributes are diversified and stochastic, 
we simultaneously build a set of category count-based index system, as shown in Figure ~\ref{Memory Architecture}.


For each layout, the count of each category can be obtained.
This is discrete and the combinations of counts of each category can be enumerated. Hence, it is suitable to be the keys in the database, in contrast to the positions or sizes that are continuous, making them uncountable.
If the types of all elements in the layout to be generated are provided, we can search for qualified layouts using the element count of each category as key, as shown on the top right of Figure ~\ref{Memory Architecture}.

However, in certain cases, attributes of only a portion of elements are known.      
In these situations, our per-category-count-based index system comes into play.
Given part of the elements as condition, we can get the lower-bound count of elements for each category.
Then for each category, the indices of layouts in which the count of the corresponding type element is greater than or equal to that lower bound can be gathered, as shown on the left side of Figure ~\ref{Memory Architecture}.
By performing intersections between layout template sets for count constraints of each category, the layout candidates that qualifiy the element count lower bound of each category can be identified.

This first-stage retrieval with type constraints helps us narrow down the range of candidate layouts and then we perform similarity measurement between query and preserved layout candidates to pick the most similar ones.
Inspired by research in dissimilarity measurement for detection evaluation~\cite{otani2022optimal} and layout structural hierarchies~\cite{manandhar2020learning, bai2023layout, DBLP:journals/corr/abs-2407-12356}, we propose to leverage bipartite matching to find the best assignment between elements in query condition and candidates and calculate similarity upon this.

We define the layout similarity as the problem of best matching of weighted bipartite graphs. 
Suppose we have query layout $l = \{e_1, e_2, ..., e_m\}$ and retrieved layout $\hat{l} = \{\hat{e_1}, \hat{e_2}, ..., \hat{e_n}\}$, where $e$ and $\hat{e}$ are the elements in the layouts, the similarity is given by 
solving this optimization problem:
\begin{equation}
\begin{aligned}
max \sum_{i=1}^m \sum_{j=1}^n w(e_i, \hat{e_j}) \cdot \gamma_{ij}, 
\ s.t. \ \gamma_{ij} \in \{0, 1\} & , \ \forall i \in \{1, 2, ..., m\}, j \in \{1, 2, ..., n\},   \\
 \ \sum_{i=1}^n \gamma_{ij} \leq 1, \ \forall i \in \{1, 2, ..., m\}, & \ \sum_{i=1}^m \gamma_{ij} \leq 1, \ \forall i \in \{1, 2, ..., n\}
\end{aligned}
\end{equation}
The $\gamma_{ij} \in \{0, 1\}$ for each pair $(e_i, \hat{e_j})$ is 1 if they are matched otherwise 0.
The weight $w(e_i, \hat{e_j})$ between between $e_i$ and $\hat{e_j}$ is defined as:
\begin{equation}
    w(e_i, \hat{e_j}) =
    \begin{cases}
        \text{IoU}(e_i, \hat{e_j}) & \text{if } T(e_i) = T(\hat{e_j}), \\
        \ \ \ \ \ \ \ 0 & \text{if } T(e_i) \neq T(\hat{e_j}),
    \end{cases}
\end{equation}
where $T(e_i)$ and $T(\hat{e_j})$ are the categories of element $e_i$ and $\hat{e_j}$ respectively.
By solving such problem with Kuhn-Munkres algorithm, we can find the best matching between elements in two layouts and obtain the similarity measurement.

When retrieving layouts, we first collect candidates using type conditions to narrow down the search space and then calculate similarities between query and type-condition-qualified candidates to identify the most similar layout.
This can uncover potential corresponding attributes beyond the initial conditions, offering more comprehensive guidance for layout generation.

\subsection{Retrieval-Augmented Layout Generation}
\label{Retrieval-Augmented Layout Generation}




Flow matching aims to estimate a flow mapping samples from a simple distribution $p_0(x)$, e.g., a Gaussian distribution to the complex target data distribution $p_1(x)$. 
Such flow is defined by a time-dependent vector field $v_t : [0, 1] \times \mathbb{R}^d \to \mathbb{R}^d$ and can be represented by the ODE:
\begin{equation}
    \frac{d}{dt} \phi_t(x) = v_t(\phi_t(x)), \quad \phi_0(x) = x_0,
\end{equation}

\begin{figure}
  \centering
  \includegraphics[width=1\textwidth]{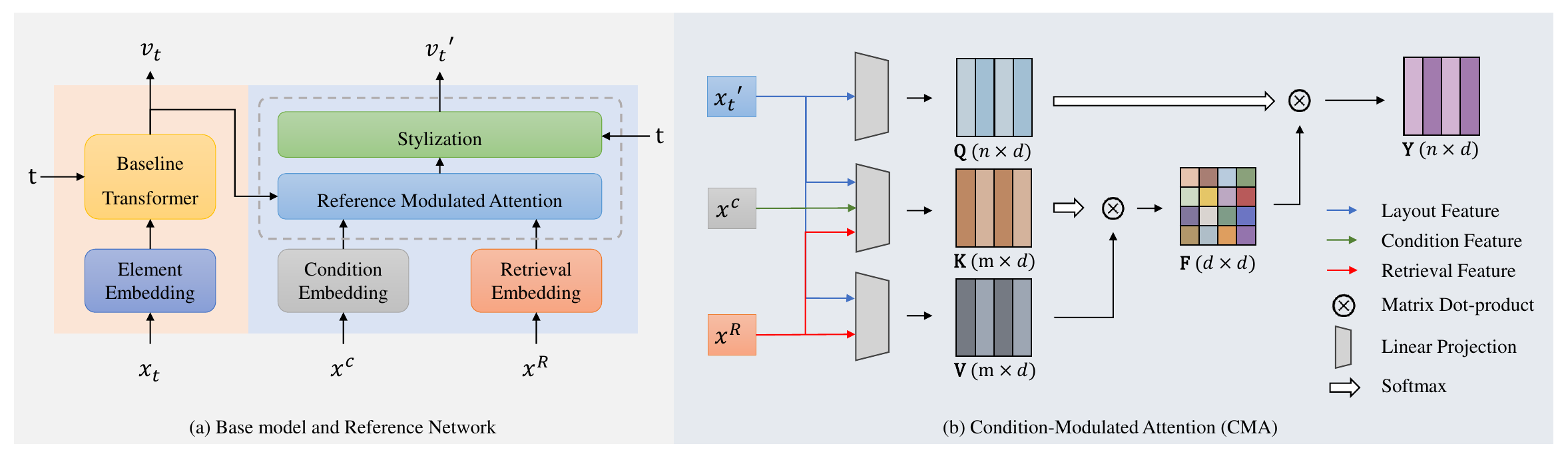}
  \caption{Retrieval Augmented Vector Field Predictor and Condition Modulated Attention.}
  \label{Reference Modulated Attention}
\end{figure}

where $\phi_t(x)$ describes how an initial sample $x_0$ is transported over time.
It trains a neural network $u_{\theta}(t, x)$ to match the vector field $v_t(x)$ through an equivalent Conditional Flow Matching objective ~\cite{lipman2023flow} 
that averts the unknown $p_t$ or $v_t$
by conditioning on a latent variable $z$:
\begin{equation}
    \mathcal{L}_{CFM}(\theta) = \mathbb{E}_{t \sim \mathcal{U}(0,1), z \sim q(z), x \sim p_t(x|z)} \| u_{\theta}(t,x) - v_t(x|z) \|^2,
\end{equation}


This allows training the model with a conditional vector field and its associated probability path and enables the model to learn the flow dynamics in a conditional setting.
In case of retrieval augmentation, a similar sample can be referenced. Hence the model can predict the vector field with the assistance of the reference.
Besides, an additional $L_1$ regulation term is added to enforce alignment between elements ~\cite{guerreiro2024layoutflow, chen2024towards}. The final training objective can be written as:
\begin{equation}
    \mathcal{L}(\theta) = \mathbb{E}_{t \sim \mathcal{U}(0,1), z \sim q(z), x \sim p_t(x|z)} \| u_{\theta}(t,x,x^R) - v_t(x|z) \|^2 + \lambda \mathcal{L}_1(\theta),
\end{equation}
where $x^R$ denotes the retrieved similar layout for reference.
Once the model is trained, we can generate new layouts by sampling from our initial distribution and solving the ODE describing the flow as defined in Eq. (3).
This can be done using any numerical ODE solver and the initial sample $x_0$ is moved step-by-step along the direction predicted by the network in an autoregressive manner as:
\begin{equation}
    \mathbf{x}_{\frac{i+1}{T}} = \mathbf{x}_{\frac{i}{T}} + \frac{1}{T} u_{\theta} \left( \frac{i}{T}, \mathbf{x}_{\frac{i}{T}}, \mathbf{x}^R \right).
\end{equation}

\textbf{Reference Network.}
Similar to previous models~\cite{inoue2023layoutdm, guerreiro2024layoutflow}, our pipeline is constructed on the foundation of transformer layers. The reference is passed into the cross-attention component, as shown in Figure ~\ref{Reference Modulated Attention}.
Unlike normal cross-attention modules, we realize the fusion of three features in
Condition Modulated Attention (CMA): the current timestep layout $x_t'$, reference layout $x^R$ and the condition $x^c$, 
as shown in Figure ~\ref{Reference Modulated Attention} (b).  
We adjust the dimensions of these three features with linear layers and fuse them through matrix dot products.
They are processed by
Linear Attention~\cite{shen2021efficient} for efficient computation.
This design enables fusing layout features from retrieved samples and also considering similarities between given conditions and corresponding attributes in the retrieved layouts.
Besides, we need to incorporate temporal information in the reference network to adjust reference feature selection in different timesteps. 
This is done 
by the scale and shift inside the stylization module after the condition-modulated attention, whose parameters are regressed from time embedding $t$.

Figure ~\ref{Reference Modulated Attention} (a) shows the baseline Transformer encoder on the left and the reference network on the right.
If reference is provided, the reference network is activated, taking the intermediate layout feature $x_t'$ from the baseline Transformer encoder, condition $x^c$ and reference $x^R$ to predict the reference-guided vector field.
Otherwise the vector field is predicted by the baseline model.



\begin{table}
  \caption{Quantitative results for various layout generation tasks on the
RICO and PubLayNet datasets. The two best results are highlighted in bold and
underlined. The → symbol indicates best results are the ones closest to the validation
data. Models marked with * have been retrained.}
  \label{performance-table}
  \centering
  \resizebox{0.9\textwidth}{!}{
  \begin{tabular}{cccccccccc}  
    \toprule
    & & \multicolumn{4}{c}{Rico} & \multicolumn{4}{c}{PubLayNet} \\
    Task & Model & FID$\downarrow$ & Ali$\rightarrow$ & Ove$\rightarrow$ & mIoU$\uparrow$ & FID$\downarrow$ & Ali$\rightarrow$ & Ove$\rightarrow$ & mIoU$\uparrow$ \\
    \midrule
    \multirow{7}{*}{C->S+P} & NDN-none & 13.76 & 0.560 & 0.550 & 0.350 & 35.67 & 0.350 & 0.170 & 0.310 \\
    & LayoutFormer++ & 2.48 & \textbf{0.124} & 0.537 & \underline{0.377} & 10.15 & 0.025 & 0.009 & 0.333 \\
    \cmidrule(r){2-10}
    & LayoutDM* & 2.39 & 0.222 & 0.598 & 0.341 & 4.20 & 0.058 & 0.030 & \textbf{0.351} \\       
    & DLT* & 6.64 & 0.303 & 0.616 & 0.326 & 7.09 & 0.097 & 0.040 & 0.349  \\
    & LayoutDiffusion & 1.56 & \textbf{0.124} & \textbf{0.491} & 0.345 & 3.73 & \textbf{0.029} & \underline{0.005} & 0.343 \\
    & LayoutFlow & \underline{1.48} & 0.176 & 0.517 & 0.322 & \textbf{3.66} & \underline{0.037} & 0.011 & \underline{0.350} \\
    & LayoutRAG (ours) & \textbf{1.22} & \underline{0.171} & \underline{0.499} & \textbf{0.388} & \underline{3.70} & \textbf{0.029} & \textbf{0.004} & 0.348  \\
    \midrule
    \multirow{4}{*}{C+S->P} & LayoutDM* & 1.76 & \underline{0.175} & 0.606 & 0.424 & 2.70 & 0.071 & 0.053 & 0.423 \\
    & DLT* & 6.27 & 0.332 & 0.609 & 0.424 & 5.35 & 0.130 & 0.053 & 0.426  \\
    & LayoutFlow & \underline{1.03} & 0.283 & \textbf{0.523} & \underline{0.470} & \underline{1.26} & \underline{0.041} & \underline{0.031} & \textbf{0.454} \\
    & LayoutRAG (ours) & \textbf{1.01} & \textbf{0.149} & \underline{0.527} & \textbf{0.489} & \textbf{0.77} & \textbf{0.040} & \textbf{0.030} & \underline{0.453} \\
    \midrule
    \multirow{3}{*}{Completion} & LayoutDM* & 5.21 & \underline{0.094} & 0.658 & 0.574 & 4.48 & 0.069 & 0.040 & 0.437 \\
    & LayoutFlow & \underline{3.59} & 0.182 & \underline{0.605} & \underline{0.628} & \underline{4.02} & \textbf{0.050} & \underline{0.024} & \underline{0.445} \\
    & LayoutRAG (ours) & \textbf{1.80} & \textbf{0.071} & \textbf{0.459} & \textbf{0.710} & \textbf{3.95} & \underline{0.051} & \textbf{0.017} & \textbf{0.458} \\
    \midrule
    \multirow{7}{*}{U-Cond} & LayoutTransformer & 24.32 & 0.037 & 0.542 & 0.587 & 30.05 & 0.067 & 0.005 & 0.359 \\
    & LayoutFormer++ & 20.20 & 0.051 & 0.546 & 0.634 & 47.08 & 0.228 & 0.001 & 0.401 \\
    \cmidrule(r){2-10}
    & LayoutDM* & 4.43 & 0.143 & 0.584 & 0.582 & 8.94 & 0.081 & 0.024 & \underline{0.427} \\
    & DLT* & 13.02 & 0.271 & 0.571 & 0.566 & 12.70 & 0.117 & 0.036 & \textbf{0.431} \\
    & LayoutDiffusion & 2.49 & \textbf{0.069} & 0.502 & \underline{0.620} & \underline{8.63} & 0.065 & \textbf{0.003} & 0.417 \\
    & LayoutFlow & \underline{2.37} & 0.150 & \underline{0.498} & 0.570 & 8.87 & \underline{0.057} & 0.009 & 0.424 \\
    & LayoutRAG (ours) & \textbf{1.96} & \underline{0.129} & \textbf{0.458} & \textbf{0.638} & \textbf{8.28} & \textbf{0.051} & \underline{0.004} & 0.425 \\
    \midrule
    & Validation Data & 2.10 & 0.093 & 0.466 & 0.658 & 8.10 & 0.022 & 0.003 & 0.434 \\
    \bottomrule
  \end{tabular}
  }
  \vspace{-0.3cm}
\end{table}

\section{Experiments}

\subsection{Experimental Setup}

\textbf{Datasets}
We evaluate our model on the RICO ~\cite{deka2017rico} and PubLayNet ~\cite{zhong2019publaynet} datasets, following previous methods. 
RICO contains over 66k User Interface (UI) layouts with 25 element types, and PubLayNet includes over 360k document layouts annotated with 5 different element types.
We train our model using the dataset split described in~\cite{jiang2023layoutformer++, zhang2023layoutdiffusion}, which discards layouts containing more than 20 elements.

\textbf{Evaluation Metrics}
We adopt four metrics to evaluate our model as previous models.
Frechet Inception Distance (\textbf{FID})~\cite{heusel2017gans} measures the overall performance by computing the distance between the distribution of the generated layouts and that of real layouts in the feature space.
We use the same network with identical weights as LayoutDiffusion~\cite{zhang2023layoutdiffusion}.
Maximum Interaction over Union (\textbf{mIoU})~\cite{kikuchi2021constrained} calculates the maximum IoU between bounding boxes of the generated layouts and those of the real layouts with the same type set to measure the similarity between real layouts and generated ones.
Alignment(\textbf{Ali})~\cite{lee2020neural} measures whether the elements in a generated layout are well-aligned, either by center or by edges.
Overlap(\textbf{Ove})~\cite{li2019layoutgan} measures the overlapping area between elements in the generated layout.

\textbf{Generation Tasks}
We evaluate our method against existing approaches on multiple layout generation tasks. 
\textbf{U-Cond} describes the layout generation task without any constraints.
\textbf{C→S+P} donates conditional generation based solely on class, and \textbf{C+S→P} represents conditional generation based on both class and size of each element. 
Then we consider the \textbf{Completion} task with attributes given for a subset of elements. 
Following LayoutDM~\cite{inoue2023layoutdm}, we randomly sample 20\% of its elements and ask the model to complement the left elements.
We do not include the refinement task~\cite{zhang2023layoutdiffusion} here because the slightly noisy layout contains valid information for layout generation, where as the retrieved samples introduce deviation upon the slightly noisy layout, making it differ more from the target layout to be generated.
Even though, our model is still able to perform refinement via 
the LayoutFlow ~\cite{guerreiro2024layoutflow} baseline branch, which has already achieved state-of-the-art performance in refinement,
as shown in the left part of Figure ~\ref{Reference Modulated Attention} (a).
For other conditional tasks, certain parts of the attributes are totally unknown. The retrieved layouts can provide valuable information for generation.

\subsection{Quantitative Analysis}
We report the results of our proposed method against state-of-the-art models on aforementioned tasks using the PubLayNet and Rico datasets in Table~\ref {performance-table}.
In terms of FID, our model outperforms state-of-the-art methods across
all four tasks, except for a close second place on the \textbf{C->S+P} task of PubLayNet,
proving its strong capabilities of retrieval augmented layout generation.
For the \textbf{C->S+P} and \textbf{Completion} tasks in Rico, LayoutRAG outperforms previous models by a large margin. This is because Rico has a relatively small amount of data with many more element types and our retrieval mechanism heavily relies on the element count of each category. Hence, more accurate layouts are retrieved to aid in generating better conditioned samples.
The \textbf{C+S->P} result on PubLayNet is also much better, indicating that our retrieval-augmented paradigm absorbs value information from references and thus produces better results.
Meanwhile, our mIoU is larger than that of other models on Rico.
Regarding the geometrical metrics Alignment and Overlap, our proposed model also produces very competitive values. 
Besides, we provide the comparison with LACE in the aligned setting in the supplemental material.

\begin{table}
  \caption{Comparison between raw retrieval results with several sota models for retrievable test data. The two best results are highlighted in bold and
underlined. Models marked with * are retrained.}
  \label{raw-retrieval-table}
  \centering
  \resizebox{0.9\textwidth}{!}{
  \begin{tabular}{cccccccccc}  
    \toprule        
    & \multirow{2}{*}{Model} & \multicolumn{2}{c}{Uncond} & \multicolumn{2}{c}{C->S+P} & \multicolumn{2}{c}{C+S->P} & \multicolumn{2}{c}{Completion} \\
    \cmidrule(r){3-4}   \cmidrule(r){5-6} \cmidrule(r){7-8} \cmidrule(r){9-10}
    & & FID$\downarrow$ & mIoU$\uparrow$ & FID$\downarrow$ & mIoU$\uparrow$ & FID$\downarrow$ & mIoU$\uparrow$ & FID$\downarrow$ & mIoU$\uparrow$ \\
    
    \midrule
    \multirow{4}{*}{Rico} & LayoutDM*  & 8.69 & \underline{0.613} & 1.93 & 0.415 & 0.96 & 0.512 & 2.69 & 0.604 \\
    & LayoutFlow & \underline{8.09} & 0.593 & \underline{1.09} & \underline{0.432} & \underline{0.68} & 0.537 & 1.95 & 0.603 \\
    & Raw Retrieval & \textbf{6.98} & \textbf{0.696} & \textbf{0.85} & \textbf{0.515} & 0.75 & \underline{0.576} & \textbf{1.27} & \textbf{0.715} \\
    & LayoutRAG & - & - & - & - & \textbf{0.55} & \textbf{0.648} & \underline{1.68} & \underline{0.704} \\
    \midrule
    \multirow{4}{*}{PubLayNet} & LayoutDM*  & 8.26 & \textbf{0.434} & 4.23 & \underline{0.353} & 2.79 & 0.425 & 4.57 & 0.434 \\
    & LayoutFlow & \underline{8.24} & \underline{0.433} & \textbf{3.75} & 0.347 & \underline{1.30} & \underline{0.448} & \underline{3.80} & 0.437 \\
    & Raw Retrieval & \textbf{8.14} & 0.414 & \underline{3.77} & \textbf{0.348} & 2.01 & 0.410 & \textbf{2.80} & \underline{0.457} \\
    & LayoutRAG & - & - & - & - & \textbf{0.76} & \textbf{0.454} & 3.94 & \textbf{0.458} \\             
    \bottomrule
  \end{tabular}
  }
  \vspace{-0.3cm}
\end{table}

\subsection{Ablation Study}

\textbf{Retrieval Result Analyses}
We provide detailed analysis of retrieval result here.
Our retrieval mechanism firstly gathers type condition-qualified layout templates and then sort them according to similarity. 
Although type-qualified templates may not be found for a small fraction of cases, in most instances, appropriate layout templates can be retrieved. 
In Rico, 66\% of the test cases can be retrieved by category-based retrieval,
and the percentage of test cases can be retrieved by category condition in PubLayNet is 99\%.
Moreover, for retrievable test cases, only 3\% of them does not have more than 20 type condition-qualified layout candidates.
This provides solid support for subsequent similarity-based retrieval.
For the completion task, only a very small part of cases (about 3\%) cannot be retrieved.
To sum up, in the vast majority of cases, a certain number of search results can be found.

Then we evaluate the retrieval results of all retrievable test data 
using the same metrics with that of generation. 
Since the aforementioned metrics are to measure similarity between two brunches of figures, they can to some extent reflect similarity between retrieval results and ground truth. The metrics for retrievable data are shown in Table ~\ref{raw-retrieval-table}.
Note the FID here is different from Table ~\ref{performance-table} because the ground truth for calculating FID is only a retrievable subset of test set.
The raw retrieval results shows great indicators, achieving similar FID with the sota generative models.
From uncondition to category-condition and then to category-size condition, the FID of Raw Retrieval group turns smaller and mIoU becomes larger, indicating the retrieval becomes more similar as more condition is given. 
This reveals that our retrieval mechanism can well utilize the given conditions to find similar templates.
The similar retrieval facilitates LayoutRAG to achieve better results with respect to previous models.
The slight drop from Raw Retrieval to LayoutRAG in Completion is because the model just modifies the attributes of 20 \% known elements to the given attributes. 
This may to some extent violate the true layout distribution, but still yeilds better performance compared with state-of-the-art models.
Therefore, retrieval is applicable in most cases and serves as an effective method for utilizing a comprehensive layout database to provide detailed conditional guidance.





\textbf{Effect of Layout Memory Database} 
To study the influence of sample quality of the database on generation results,
we experiment to see the generation results guided by references with different iou-based similarities.
We discard retrievals with similarities larger than certain thresholds to simulate different reference qualities.
This experiment is conducted using \textbf{C+S->P} task on PubLayNet, shown in Table~\ref{memory-ablation}.
When the reference is not similar, the generation quality is close to the base model, LayoutFlow. This is because our model is built on pre-trained LayoutFlow, and by feeding irrelevant references with a certain probability during training, the model learns to rely on the base model instead of references when the references are dissimilar.
As the similarity threshold of discarding increases, the best retrievals become more similar, and the generation quality improves to approach our state-of-the-art standard.
This observation suggests that by having more relevant samples in the database, the model's behavior can be tailored accordingly. As a result, the proposed method enables a flexible way to remedy failure cases without additional training or fine-tuning.


\begin{table}[h!]
    \centering
    \begin{minipage}{0.45\textwidth}
        \centering
        \caption{Generation results with different reference sample qualities.}   
        \label{memory-ablation}
        \resizebox{0.9\textwidth}{!}{
        \begin{tabular}{ccccc}
            \toprule
            Threshold & FID$\downarrow$ & Ali$\rightarrow$ & Ove$\rightarrow$ & mIoU$\uparrow$ \\
            \midrule
            0.1 & 1.24 & 0.043 & 0.039 & 0.448 \\
            0.2 & 1.15 & 0.043 & 0.036 & 0.446 \\
            0.3 & 0.99 & 0.043 & 0.034 & 0.450 \\
            0.4 & 0.84 & 0.042 & 0.031 & 0.453 \\
            0.5 & 0.81 & 0.041 & 0.031 & 0.452 \\
            0.6 & 0.79 & 0.040 & 0.031 & 0.452 \\
            \midrule
            Preserve All & 0.77 & 0.040 & 0.030 & 0.453 \\       
            \bottomrule
        \end{tabular}
        }
    \end{minipage}\hfill
    \begin{minipage}{0.55\textwidth}
        \centering
        \caption{Performance comparison for different approaches of fusing retrieval features into generation.}
        \label{rma-ablation}
        \resizebox{0.9\textwidth}{!}{
        \begin{tabular}{ccccc}
            \toprule
            & FID$\downarrow$ & Ali$\rightarrow$ & Ove$\rightarrow$ & mIoU$\uparrow$ \\
            \midrule
            LayoutFlow Base & 1.26 & 0.041 & 0.031 & 0.454 \\
            Concat + Linear & 1.20 & 0.042 & 0.037 & 0.446 \\
            Vanilla Cross Attn & 0.90 & 0.042 & 0.032 & 0.448 \\
            CMA (w/o cond) & 0.83 & 0.040 & 0.030 & 0.450 \\
            \midrule
            CMA & 0.77 & 0.040 & 0.030 & 0.453 \\
            \bottomrule
        \end{tabular}
        }
    \end{minipage}
    \vspace{-0.2cm}
\end{table}

\textbf{CMA Analyses} 
We further evaluate the component of integrating retrieval as reference guidance in the generation process.
We conduct this experiment on \textbf{C+S->P} task of PubLayNet and use the LayoutFlow as the baseline model. 
The Condition-Modulated Attention module shown on the right of Figure~\ref{Reference Modulated Attention} is replaced by different fusion models to see the effect.
We compare our model with the versions of fusion by linear layer, vanilla cross attention, and an CMA variant without condition embedding.
The results are shown in Table ~\ref{rma-ablation}.
The experiments reveal that compared to the basic cross-attention, our method can better absorb valid information from retrieval reference to generate high-quality layouts.
The additional condition embedding enables the model to adaptively fuse retrieval layout information considering condition similarity and difference.
By contrast, linear-based methods concatenate the inputs and references, which hinders the direct extraction of meaningful information from the references, leading to relatively modest performance.

\begin{figure}
  \centering
  \includegraphics[width=1\textwidth]{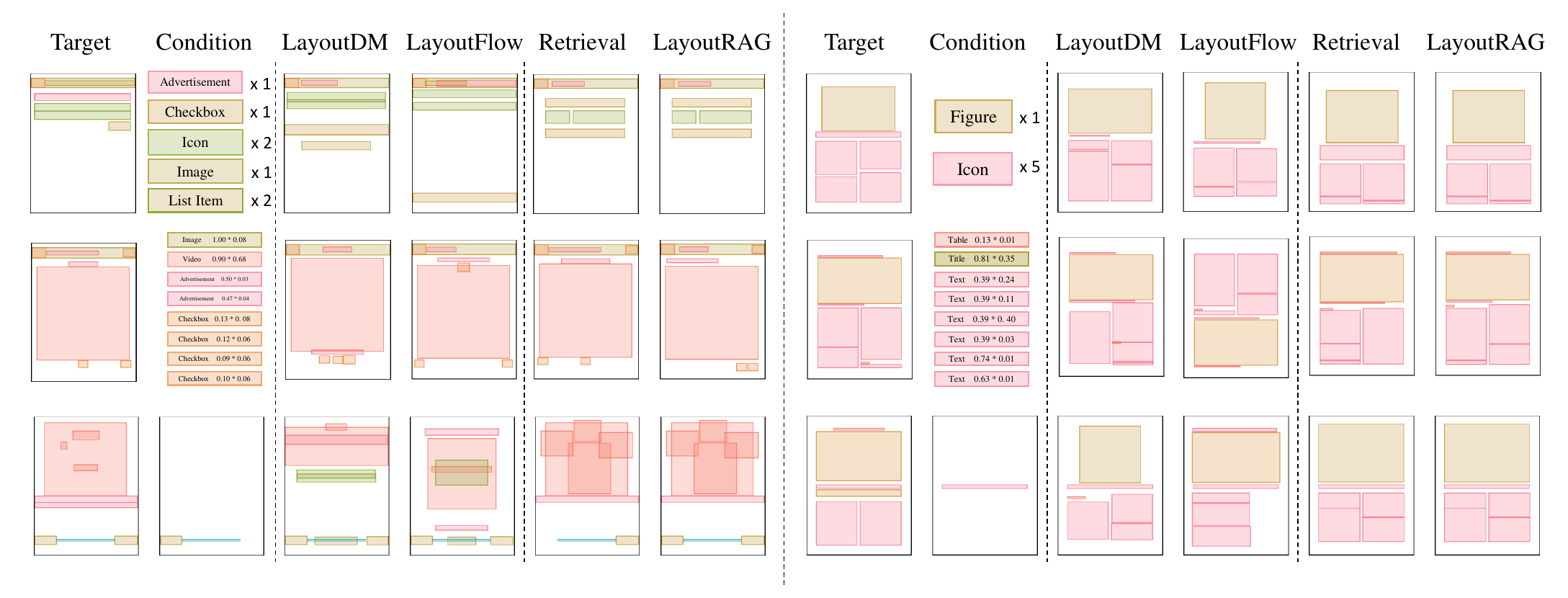}
  \caption{Visualization of conditional generation results on Rico(left) and PubLayNet(right). The three rows show the results for \textbf{C->S+P}, \textbf{C+S->P}, and \textbf{Completion} respectively.}
  \label{Visualization}
  \vspace{-0.2cm}
\end{figure}

\subsection{Qualitative Evaluation}
We provide some qualitative examples in Figure ~\ref{Visualization}. 
More samples can be found in the supplemental material. 
Overall, LayoutRAG shows a strong performance
across the conditional tasks, producing visually pleasing results that resemble real layouts.

\section{Conclusion}
In this paper, we propose to implement conditional layout generation by reference layout retrieval and reference-guided generation.
Experiments show that our model outperforms state-of-the-art models.
Through retrieving in term of known layout attributes, the model gets more aware of potential arrangements of unknown attributes, and thus produces better conditional layouts.
In addition, retrieval augmentation provides flexibility to tailor the behavior of the model after training in term of relevant samples in the database.
Overall, our model provides a novel conditioning formulation that enrichs the condition through retrieval to make more comprehensive guidance and enables more flexibility for model poct-hoc modification.
{\small \bibliographystyle{plainnat} \bibliography{references}}





\end{document}